\pdfoutput=1
\PassOptionsToPackage{prologue,dvipsnames}{xcolor}
\documentclass[11pt]{article}

\usepackage[preprint]{acl}

\usepackage{times}
\usepackage{latexsym}
\usepackage{array}
\usepackage{graphicx}
\usepackage{epstopdf}
\usepackage{makecell}
\newcolumntype{L}[1]{>{\raggedright\let\newline\\\arraybackslash\hspace{0pt}}m{#1}}
\newcolumntype{C}[1]{>{\centering\let\newline\\\arraybackslash\hspace{0pt}}m{#1}}
\newcolumntype{R}[1]{>{\raggedleft\let\newline\\\arraybackslash\hspace{0pt}}m{#1}}
\usepackage{multirow}
\usepackage{svg}
\usepackage{makecell}
\usepackage[utf8]{inputenc}
\usepackage{amsmath}
\usepackage{cleveref}
\crefname{section}{§}{§§}
\Crefname{section}{§}{§§}
\usepackage{booktabs}
\usepackage[dvipsnames]{xcolor}
\usepackage{colortbl}
\usepackage{color}
\usepackage{algorithm}
\usepackage{algorithmic}
\usepackage{pifont}
\usepackage{arydshln}
\hypersetup{
    colorlinks=true,
    linkcolor=blue,
    filecolor=blue,      
    urlcolor=blue,
    citecolor=cyan,
}
\usepackage{amssymb}
\usepackage{lipsum}
\usepackage{tcolorbox}

\usepackage[T1]{fontenc}

\usepackage[utf8]{inputenc}

\usepackage{microtype}

\usepackage{inconsolata}

\usepackage{graphicx}

\title{DATA: Decomposed Attention-based Task Adaptation for\\ Rehearsal-Free Continual Learning}
\author{Huanxuan Liao$^{1,2}$, Shizhu He$^{1,2}$\thanks{Corresponding author}, Yupu Hao$^{1,2}$, \textbf{Jun Zhao}$^{1,2}$, \textbf{Kang Liu}$^{1,2}$  \\
    $^1$ The Key Laboratory of Cognition and Decision Intelligence for Complex Systems, \\
    Institute of Automation, Chinese Academy of Sciences, Beijing, China \\
    $^2$ School of Artificial Intelligence, University of Chinese Academy of Sciences, Beijing, China \\
  {\{liaohuanxuan2023, haoyupu2023\}@ia.ac.cn} {\{shizhu.he, jzhao, kliu\}@nlpr.ia.ac.cn} \\}

\begin{document}
\maketitle
\begin{abstract}
Continual learning (CL) is essential for Large Language Models (LLMs) to adapt to evolving real-world demands, yet they are susceptible to catastrophic forgetting (CF).  While traditional CF solutions rely on expensive data rehearsal, recent rehearsal-free methods employ model-based and regularization-based strategies to address this issue.
However, these approaches often neglect the model's plasticity, which is crucial to achieving optimal performance on newly learned tasks. Consequently, a key challenge in CL is striking a balance between preserving plasticity and mitigating CF.  
To tackle this challenge, we propose the \textbf{D}ecomposed \textbf{A}ttention-based \textbf{T}ask \textbf{A}daptation (DATA), which explicitly decouples and learns both task-specific and task-shared knowledge using high-rank and low-rank task adapters (e.g., LoRAs). For new tasks, DATA dynamically adjusts the weights of adapters of different ranks based on their relevance and distinction from previous tasks, allowing the model to acquire new task-specific skills while effectively retaining previously learned knowledge.
Specifically, we implement a decomposed component weighting strategy comprising learnable components that collectively generate attention-based weights, allowing the model to integrate and utilize diverse knowledge from each DATA. Extensive experiments on three widely used benchmarks demonstrate that our proposed method achieves state-of-the-art performance. Notably, our approach significantly enhances model plasticity and mitigates CF by extending learnable components and employing stochastic restoration during training iterations. 
Our code will be available at \url{https://github.com/Xnhyacinth/DATA}.

\end{abstract}

\section{Introduction}

\begin{figure}[t]
\centerline{\includegraphics[width=0.5\textwidth]{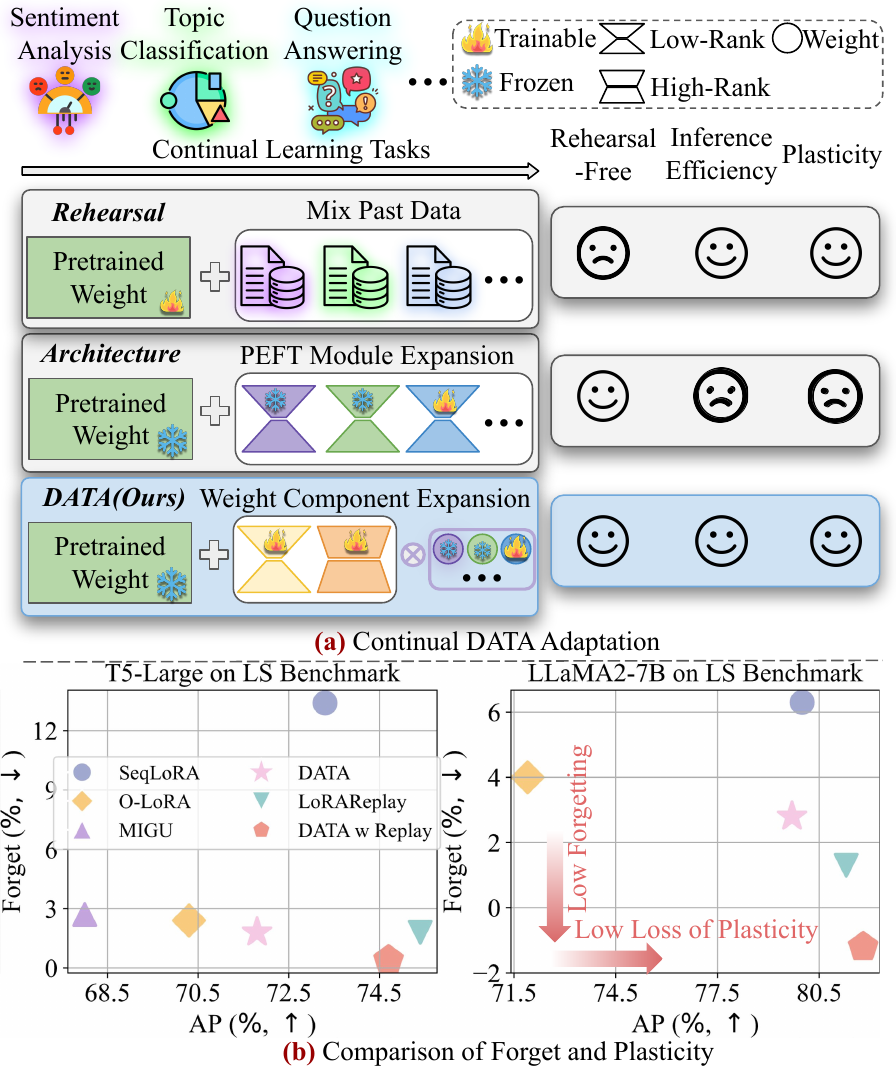}}
\caption{
\textcolor{red}{(a)} Comparisons of previous CL methods with \textbf{DATA}. \textit{Rehearsal-free} indicates that the methods do not require storing sample features from previous tasks. \textit{Inference Efficiency} denotes the computational efficiency during the inference phase. \textit{Plasticity} is the ability to adapt effectively to new tasks.
\textcolor{red}{(b)} Comparison of \textbf{AP} and \textbf{Forget} (Sec. \ref{sec:dataset}) across different CL methods.
}
\label{intro}
\end{figure}

Continual learning (CL) \cite{orthogonal, clsurvey} is essential for Large Language Models (LLMs) \cite{llama3, Qwen2TR} to continuously evole, adapting to real-world demands and progressively improve regarding a series of new tasks. Unlike traditional supervised learning, which relies on independent and identically distributed (i.i.d.) data, CL focuses on the dynamic and evolving demands of real-world applications \cite{investigating, continualsurvey}, such as domain-specific adaptations \cite{codellama} and alignment with human preferences \cite{rlhf}.
Departing from the i.i.d. assumption presents two major challenges: 1) \textbf{Catastrophic Forgetting} (CF) --- the tendency of a model to overwrite previously learned knowledge when acquiring new tasks \cite{1989Catastrophic}. 2) \textbf{Loss of Plasticity}---the model’s diminishing ability to adapt effectively to distribution shifts over time \cite{Dohare2021ContinualBS}.


As illustrated in Figure \ref{intro} (a), an ideal CL method leveraging LLMs should possess three essential properties. A commonly used strategy in CL is the preservation \cite{inscl} or synthesis \cite{Sun2019LAMOLLM} of past training data, referred to as \textbf{rehearsal} (or \textit{replay}). However, this approach presents notable challenges, particularly in scenarios involving \textit{sensitive user data} where long-term storage is impractical \cite{llama}. Furthermore, rehearsal methods require \textit{increased storage capacity and computational resources} during the learning process \cite{fine}. Architecture-based methods \cite{Wang2024RehearsalFreeMA} aim to mitigate interference between new and previously learned tasks by dynamically expanding the model's capacity or isolating existing model weights. Nevertheless, these approaches often necessitate training separate expert models for each task, which in turn requires selecting \cite{sapt} or combining \cite{olora} modules during testing. This can limit their ability to generalize effectively to long sequence tasks.
Moreover, as shown in Figure \ref{intro} (b), these methods mitigate CF by \textit{sacrificing new task accuracy} (i.e., insufficient \textit{plasticity}), prompting us to explore innovative perspectives for enhancing CL capacity.


To address these limitations while maintaining plasticity and mitigating CF, we propose a \textbf{D}ecomposed \textbf{A}ttention-based \textbf{T}ask \textbf{A}daptation (DATA), as depicted in Figure \ref{intro} (a), which explicitly manages high-rank and low-rank adapters for task-specific and task-shared knowledge during the CL. For new tasks, DATA dynamically adjusts the weights of different ranks adapters based on their relevance and distinction from previous tasks, enabling the model to acquire new task-specific skills while preserving previously learned capabilities. 
To extract knowledge across tasks more effectively, we introduce a \textit{decomposed component weighting} strategy comprising learnable components that jointly generate attention-based weights, allowing dynamic fusion of knowledge from each rank DATA. 
During inference, low-rank and high-rank DATA can be reparameterized and projected into LLMs without increasing parameters, preserving the model plasticity \cite{vida}. We use \textit{stochastic restoration} \cite{cotta, dohare2024loss} technique to further protect source knowledge and mitigate CF.

We conduct extensive experiments to evaluate DATA on Standard CL Benchmark \cite{cl}, Long Sequence Benchmark \cite{pro} and TRACE \cite{trace} across T5 model \cite{t5}, Llama2 \cite{llama}, Llama3.1 \cite{llama3} and Qwen2.5 \cite{Qwen2TR}. DATA achieves state-of-the-art performance on both public benchmarks and in generalizing to unseen tasks. Notably, DATA can be integrated with existing mainstream methods to further enhance the capabilities of CL.
In summary, our contributions are as follows:
\begin{itemize}
    \item We propose a method DATA for rehearsal-free CL of LLMs that balances plasticity and mitigates CF by explicitly managing both \textit{task-specific} and \textit{task-shared} knowledge.
    \item We explore different task representations of high- and low-rank adapters, decomposing knowledge into shared and specific components. This approach allows for better modeling of the relevance and uniqueness of new tasks concerning previous ones.
    \item To address various distribution shifts for each target sample, we introduce a \textit{decomposed component weighting} strategy for DATA, which dynamically fuses knowledge from low-rank and high-rank adapters, enhancing their task representations.
    \item Experiments on three benchmarks show that our approach outperforms most state-of-the-art techniques, significantly alleviating CF and improving model plasticity.
\end{itemize}

\section{Related Work}

\noindent \textbf{Continual Learning (CL).} CL aims to effectively acquire new task knowledge during ongoing training while preserving knowledge from previously learned tasks. Traditional CL methods can be broadly classified into three categories: \textit{rehearsal-based}, \textit{regularization-based}, and \textit{architecture-based} approaches. Rehearsal-based methods mitigate catastrophic forgetting (CF) by selectively retaining samples \cite{GCRGC, inscl, seekr} or pseudo-generative examples \cite{ContinualLW, Sun2019LAMOLLM} from prior tasks. In contrast, regularization-based methods employ quadratic regularization terms to constrain the update of weights critical to previous tasks \cite{Zhu2024ModelTM, migu}, thereby stabilizing learned knowledge. Architecture-based methods introduce extra task-specific parameters for each new task \cite{olora, sapt}, ensuring that previously acquired knowledge remains intact. These methods typically rely on access to old task data or involve discrete component learning, which presents significant challenges for the CL. In contrast, DATA leverages representations from diverse rank spaces and harnesses the inherent capabilities of LLMs to enable more effective CL \cite{cotta}.

\begin{figure*}[t]
\centerline{\includegraphics[width=0.95\textwidth]{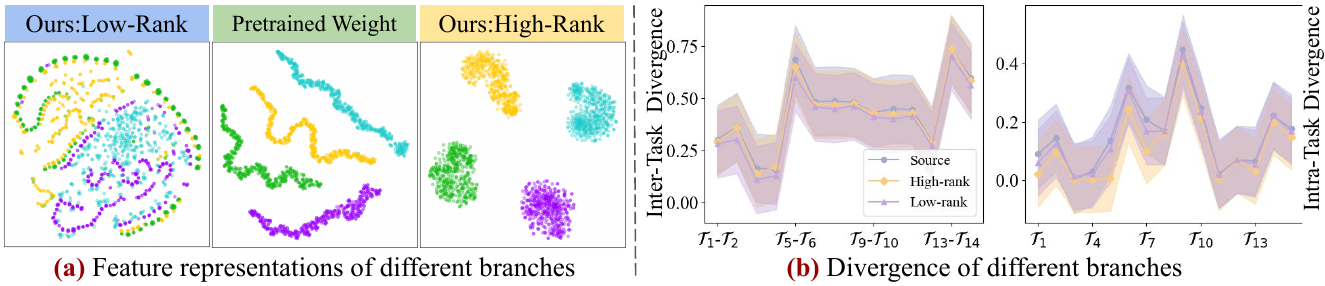}}
\caption{\textcolor{red}{(a)} We perform a t-SNE distribution analysis of different adapter representations on Order 1(4 tasks). The low-rank branch shows a consistent distribution across the target tasks and the high-rank branch exhibits substantial distribution differences across the target tasks. \textcolor{red}{(b)} We calculate the divergence of different branches in Order 4 (15 tasks). In comparison to the source model, low-rank adapters effectively alleviates inter-task divergence across all 14 task transitions, while the high-rank adapters significantly enhances intra-task feature aggregation.
}
\label{div}
\end{figure*}

\noindent \textbf{Parameter Efficient Fine-Tuning (PEFT).} LLMs have shown remarkable adaptability across a wide range of downstream tasks. However, traditional full fine-tuning for each task can incur \textit{significant computational and storage overheads}, often resulting in overfitting \cite{slora}. To mitigate these challenges, researchers have increasingly turned to PEFT methods, which can achieve comparable or even superior performance and generalization by fine-tuning only a small subset of parameters.
For instance, \textit{adapters} \cite{adapter} incorporate additional modules into various layers of the Transformer architecture, while \textit{prompt-tuning} \cite{prompt} and \textit{prefix-tuning} \cite{prefix} introduce learnable soft tokens at different layers of the Transformer input. \textit{Low-rank adaptation} (LoRA) \cite{hu2022lora} inserts low-rank branches into pre-trained weights and fine-tunes only those branches. Although these common PEFT techniques offer certain advantages, they are generally confined to static single-task or multi-task learning scenarios. In contrast, DATA leveraging the effectiveness and efficiency of LoRA enables continual adaptation by dynamically combining LoRAs of two different ranks \cite{vida} utilizing attention-based weights.


\section{Methodology}

\noindent \textbf{Preliminary.} Continual learning (CL) seeks to tackle challenges within ongoing sequences, specifically addressing how to adapt to new tasks without forgetting previously learned knowledge. Formally, suppose there are $\mathcal{N}$ sequential tasks $ \{\mathcal{T}_1, \mathcal{T}_2, ..., \mathcal{T}_\mathcal{N}\} $, where each task $ \mathcal{T}_i = \{ \boldsymbol{x}_n^{i}, \boldsymbol{y}_n^{i} \}_{n=1}^{\mathcal{N}_i} $ consists of $\mathcal{N}_i$  training examples, with $\boldsymbol{x}_n^{i}$ representing the input and $\boldsymbol{y}_n^{i}$ is corresponding label. Let $ \mathcal{L}_i(\cdot) $ denote the empirical risk on the \( i \)-th task $ \mathcal{T}_i $, and \( f_{\theta} \) represent the model with parameters \( \theta \). The goal is for the model to perform well on both the current task \( \mathcal{T}_j \) and all previously learned tasks \( \mathcal{T}_i \), where \( i \in \{1, 2, \dots, j-1\} \).
The objective function of CL can be expressed as:
\begin{equation}
    \mathcal{L}_{\text{CL}}(\theta) = \frac{1}{j} \sum_{i=1}^{j} \mathcal{L}_i(\theta)
\end{equation}
where \( j \) is the index of the current training task.



\subsection{Motivation}
\label{moti}
Recent advances have highlighted the effectiveness of using adapters with varying ranks to model different tasks, offering a promising solution to the challenges of CL. This motivates us to further explore and validate the core principles behind utilizing both low-rank and high-rank adapters in CL.

To further explore this phenomenon, we conduct a t-SNE distribution analysis \cite{tsne} to examine the feature distributions of Order1 (4 tasks, Sec. \ref{sec:dataset}) within the 11th transformer block of LLaMA2-7B. The results as presented in Figure \ref{div} (a) reveal that low-rank adapters maintain relatively consistent feature distributions across different target tasks. This consistency suggests that the low-rank embedding space effectively mitigates the effects of dynamic distribution shifts by concentrating on extracting task-shared knowledge. Conversely, high-rank adapters display noticeable variability across tasks. This variability underscores the enhanced ability of high-rank adapters to aggregate features within a single task, highlighting their suitability for capturing the unique data distribution of each target task. 

Furthermore, we utilize the \textit{H-divergence} metric to assess the representation consistency of adapters on Order 4 (15 tasks). A relatively small inter-task divergence indicates that feature representations remain stable and are less affected by cross-task shifts \cite{Ganin2015DomainAdversarialTO}, while a small intra-task divergence within a given task indicates that the model better understands the current distribution. We compare the divergence values obtained from the source model alone, injecting low-rank adapters and injecting high-rank adapters, as shown in Figure \ref{div}). Compared to the source model and high-rank adapters, low-rank adapters produce feature representations with lower inter-task divergence, particularly when addressing mid-range target tasks or significant task shifts between adjacent tasks (i.e., tasks 6-10). These findings underscore the efficacy of low-rank adapters in acquiring long-term, task-shared knowledge during the CL.
Furthermore, high-rank adapters effectively reduce intra-task divergence across nearly all tasks, indicating their superior capability in adapting to the current task distribution and extracting task-specific knowledge in consecutive target tasks.

In summary, the structure of low-rank DATA reduces feature redundancy, placing the model in an underfitting state and enhancing its plasticity. This design enables the model to acquire general information across consecutive target tasks and effectively extract task-shared knowledge. In contrast, high-rank DATA aligns more closely with the target data distribution, allowing it to focus on learning task-specific knowledge and mitigating CF.


\begin{figure*}[t]
\centerline{\includegraphics[width=1.0\textwidth]{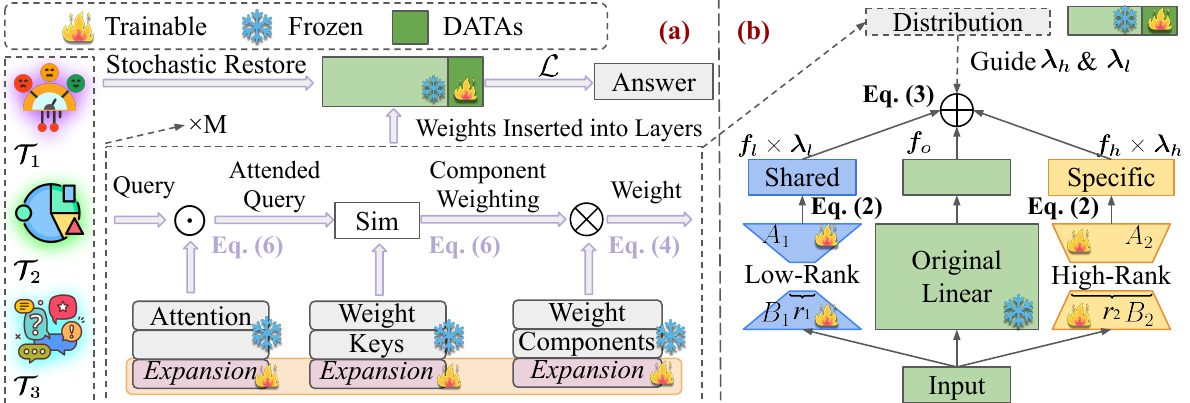}}
\caption{\textbf{Overview of the DATA framework.} \textcolor{red}{(a)} We introduce a novel \textit{decomposed component weighting} strategy for generating \textbf{attention-based weights}, parameterized by a set of \textit{extended} weight components, each associated with a corresponding key and attention vector. Only the weighting and adapter parameters are trainable which is \textit{parameter efficient} and no training data is stored for the replay which is \textit{memory-efficient} and \textit{privacy-preserving}.
\textcolor{red}{(b)} We integrate low-rank and high-rank DATA into the linear layers of the pre-trained model guided by the generated weights, allowing for the dynamic fusion of knowledge from each DATA with different task representations.
}
\label{model}
\end{figure*}

\subsection{High- and Low-Rank Adapter}
The above observations prompted us to introduce high- and low-rank Adapters into the model, aiming to simultaneously adapt current task distribution while maintaining task-shared knowledge.

Similar to LoRA, DATA's design principle is simple and effective, as illustrated in Figure \ref{model} (b). The architecture consists of three sub-branches: the central branch, which originates from the original model's linear layer, and the right and left branches, which are bottleneck structures representing high-rank and low-rank DATA, respectively.
Specifically, the right branch (high-rank) consists of an up-projection layer with parameters $ \mathbf{A}_2 \in \mathbb{R}^{d \times d_h} $, and a down-projection layer with parameters \( \mathbf{B}_2 \in \mathbb{R}^{d_h \times d} \), where \( d_h \) (e.g., \( d_h = 32 \)) represents the middle dimension of the high-rank feature.
The same principle applies to the low-rank branch, except it satisfies \( d_l < d_h \).  
For an input $\boldsymbol{x}$, the resulting high-rank features \( \boldsymbol{f}_h \) and low-rank features \( \boldsymbol{f}_l \) from the DATA are formulated as:
\begin{equation}
    \boldsymbol{f}_h = \mathbf{A}_2 \cdot (\mathbf{B}_2 \cdot \boldsymbol{x}); \boldsymbol{f}_l = \mathbf{A}_1 \cdot (\mathbf{B}_1 \cdot \boldsymbol{x})
\end{equation}

The two-branch bottleneck is connected to the output feature of the original linear \( \boldsymbol{f}_o \) through a residual connection, with scale weights \( \boldsymbol{\lambda}_h \) and \( \boldsymbol{\lambda}_l \). The fusion of knowledge \( \boldsymbol{f_x} \) is then expressed as:
\begin{equation}
    \boldsymbol{f_x} = \boldsymbol{f}_o + \boldsymbol{\lambda}_h \times \boldsymbol{f}_h + \boldsymbol{\lambda}_l \times \boldsymbol{f}_l
\end{equation}
The task knowledge scale weights \( \boldsymbol{\lambda}_h \) and \( \boldsymbol{\lambda}_l \) can be adaptively computed through \textit{decomposed component weighting} strategy as detailed in Sec. \ref{weigth}. During the inference process, the distinct task representations of the DATA are reparameterized and projected into the original model, enabling dynamic knowledge adjustments. 

\begin{table*}[t]
  \centering
  \resizebox{\linewidth}{!}{
      \begin{tabular}{L{1em}lccccccccc}
        \toprule
         & \multirow{2}{*}{\textbf{Methods}} & \multicolumn{3}{c}{\textbf{Standard CL Benchmark (SC)}}& \multicolumn{3}{c}{\textbf{Long Sequence Benchmark (LS)}} & \multicolumn{3}{c}{\textbf{TRACE}}\\
        \cmidrule(r){3-5}\cmidrule(r){6-8}\cmidrule(r){9-11}
         & & \textbf{FP} $\uparrow$ & \textbf{AP} $\uparrow$ & \textbf{Forget} $\downarrow$ & \textbf{FP} $\uparrow$ & \textbf{AP} $\uparrow$ & \textbf{Forget} $\downarrow$ & \textbf{FP} $\uparrow$ & \textbf{AP} $\uparrow$ & \textbf{Forget} $\downarrow$  \\
        \midrule
         \multirow{6}*{\rotatebox{90}{T5-Large}} & SeqLoRA        & 70.7$_{\pm .39}$ & \textbf{76.7}$_{\pm .43}$ & 6.0 & 59.9$_{\pm .56}$ & 73.3$_{\pm .28}$ & 13.4 & 12.1$_{\pm .82}$ &  44.5$_{\pm .94}$ & 32.4 \\
        & LoRAReplay  & 73.3$_{\pm .42}$ & 76.6$_{\pm .51}$ & 3.3 & 73.6$_{\pm .36}$ & \textbf{75.4}$_{\pm .59}$ & 1.8 & 34.0$_{\pm .62}$ &  \textbf{46.8}$_{\pm .63}$ & 12.8\\
        & O-LoRA \cite{olora} & 72.0$_{\pm .63}$ & 74.4$_{\pm .47}$ & 2.4 & 67.9$_{\pm .82}$ & 70.3$_{\pm .65}$ & 2.4 & - & - & - \\
        & ~~~ + MIGU \cite{migu} & 71.6$_{\pm .45}$ & 73.9$_{\pm .67}$ & 2.3 & 65.3$_{\pm .35}$ & 68.0$_{\pm .47}$ & 2.7 & - & - & - \\
        & \textbf{DATA} (\textit{ours}) & 72.7$_{\pm .58}$ & 74.8$_{\pm .68}$ & 2.1 & 70.0$_{\pm .44}$ & 71.8$_{\pm .36}$ & 1.8 & 16.7$_{\pm .21}$ &  41.3$_{\pm .58}$ & 24.6\\
        & ~~~ + Replay & \textbf{73.9}$_{\pm .24}$ & 74.5$_{\pm .75}$ & \textbf{0.6} & \textbf{74.3}$_{\pm .35}$ & 74.7$_{\pm .91}$ & \textbf{0.4} & \textbf{36.5}$_{\pm .32}$ &  45.2$_{\pm .61}$ & \textbf{8.7}\\
        \midrule
        \multirow{5}*{\rotatebox{90}{LLaMA2-7B}} & SeqLoRA        & 74.9$_{\pm .42}$ & \textbf{80.7}$_{\pm .38}$ & 5.8 & 73.7$_{\pm .66}$ & 80.0$_{\pm .65}$ & 6.3 & 64.1$_{\pm .49}$ & 77.9$_{\pm .54}$ & 13.8 \\
        & LoRAReplay  & 79.2$_{\pm .53}$ & \textbf{80.7}$_{\pm .61}$ & 1.5 & 80.0$_{\pm .45}$ & \textbf{81.3}$_{\pm .57}$  & 1.3 & 71.9$_{\pm .62}$ & \textbf{78.6}$_{\pm .75}$ & 6.7 \\
        & O-LoRA \cite{olora} & 76.4$_{\pm .74}$ & 78.7$_{\pm .59}$ & 2.3 & 67.9$_{\pm .74}$ & 71.9$_{\pm .49}$ & 4.0 & 35.0$_{\pm .34}$ & 47.0$_{\pm .42}$ & 12.0 \\
        & \textbf{DATA} (\textit{ours}) & 79.8$_{\pm .54}$ & 80.3$_{\pm .49}$ & 0.5 & 76.9$_{\pm .35}$ & 79.7$_{\pm .36}$ & 2.8  & 66.0$_{\pm .33}$ & 74.6$_{\pm .38}$ & 8.6 \\
        & ~~~ + Replay & \textbf{80.0}$_{\pm .24}$ & 80.4$_{\pm .45}$ & \textbf{0.4} & \textbf{81.8}$_{\pm .61}$ & 80.6$_{\pm .84}$ & \textbf{-1.2} & \textbf{73.1}$_{\pm .46}$ & 77.5$_{\pm .98}$ & \textbf{4.4}  \\
        \midrule
        \multirow{5}*{\rotatebox{90}{LLaMA3.1-8B}} & SeqLoRA        & 79.6$_{\pm .62}$ & 80.8$_{\pm .49}$ & 5.8 & 74.8$_{\pm .58}$ & 83.8$_{\pm .52}$ & 9.0 & 65.1$_{\pm .69}$ & 82.4$_{\pm .54}$ & 17.3 \\
        & LoRAReplay  & 80.3$_{\pm .71}$ & \textbf{80.9}$_{\pm .56}$ & 0.6 & 82.0$_{\pm .69}$ & \textbf{85.0}$_{\pm .75}$  & 3.0 & \textbf{78.7}$_{\pm .83}$ & \textbf{85.7}$_{\pm .68}$ & 7.0 \\
        & O-LoRA \cite{olora} & 72.3$_{\pm .86}$ & 73.9$_{\pm .68}$ & 1.6 & 71.4$_{\pm .64}$ & 74.8$_{\pm .64}$ & 3.7 & 36.7$_{\pm .57}$ & 50.1$_{\pm .37}$ & 13.4 \\
        & \textbf{DATA} (\textit{ours}) & \textbf{80.9}$_{\pm .40}$ &  80.6$_{\pm .35}$ & -0.3  & 80.0$_{\pm .53}$ & 82.3$_{\pm .48}$ & 2.3  & 72.7$_{\pm .94}$ & 80.4$_{\pm .88}$ & 7.7 \\
        & ~~~ + Replay & 80.8$_{\pm .33}$ & 80.4$_{\pm .47}$ & \textbf{-0.4} & \textbf{82.2}$_{\pm .66}$ & 82.6$_{\pm .77}$ & \textbf{0.4} & 77.6$_{\pm .37}$ & 81.0$_{\pm .72}$ & \textbf{3.4} \\
        \midrule
      \end{tabular}
      }
    \caption{Performance of baselines and ours \textbf{DATA} on standard CL benchmark (Order 1,2,3) and long sequence benchmark (Order 4,5,6) and TRACE (Order 7). \textbf{Bold} indicates the best in each setting. We report the mean and standard deviation of results with 3 different runs.
    }
    \label{ms}
\end{table*}

\subsection{Decomposed Component Weighting}
\label{weigth}
To effectively mitigate CF across various tasks and samples, it is crucial to extract and manage different types of knowledge. While the distinct structure of low-rank and high-rank DATA facilitates learning diverse task representations, the continual adaptation process also necessitates the normalization of knowledge fusion weights. This normalization ensures the effective capture of relevant task-specific knowledge while preserving long-term task-specific knowledge.

As illustrated in Figure \ref{model} (a), we draw inspiration from \cite{CODAPromptCD} and propose a \textit{decomposed component weighting} strategy. Specifically, we introduce a set of weight components that form a decomposed weight through weighted summation, which is then passed to the corresponding DATA layer. Furthermore, in new tasks, the weights will essentially reuse previously acquired knowledge about past tasks instead of initializing new task weights from scratch:
\begin{equation}
    \boldsymbol{\lambda} = \sum_m \boldsymbol{\alpha}_m \mathbf{W}_m
\end{equation}
where \( W \in \mathbb{R}^{L_w \times d \times M} \) represents our set of weight components, with \( M \) denoting the length of the set (i.e., the extra axis of additional capacity), $L_w$ is the weight length (chosen as a hyperparameter) and \( \boldsymbol{\alpha} \) is a \textit{weighting vector} that determines the contribution of each component, which is the cosine similarity $\gamma$ between a query \( q(\boldsymbol{x}) \) and keys:
\begin{equation}
    \boldsymbol{\alpha} = \gamma (q(\boldsymbol{x}), \mathbf{K})
\end{equation}
where \( \mathbf{K} \in \mathbb{R}^{d \times M} = \{ \mathbf{K}_1, \mathbf{K}_2, \dots, \mathbf{K}_M \} \) contains keys corresponding to weight components. The intuition behind this is that the contribution of each weight component \( \mathbf{W}_m \) to the final weight \( \boldsymbol{\lambda} \) is weighted according to the similarity.

To address the challenge of query matching, we introduce an additional component: \textit{attention}. Alongside \( \mathbf{K}_m \), each weight component \( \mathbf{W}_m \) is paired with a corresponding attention vector \( \mathbf{A}_m \), enabling the query to focus on specific features within the high-dimensional query. We utilize a simple feature selection attention mechanism, where the query vector is element-wise multiplied by the attention vector to generate an \textit{attended query} which is then used to compute the similarity with the key. Specifically, our updated approach to producing the weighting vector is as follows:
\begin{equation}
    \boldsymbol{\alpha} = \gamma(q(\boldsymbol{x}) \odot \mathbf{A}, \mathbf{K})
\end{equation}
where \( \mathbf{A} \in \mathbb{R}^{d \times M} = \{ \mathbf{A}_1, \mathbf{A}_2, \dots, \mathbf{A}_M \} \) represents the learnable attention vectors corresponding to each weight component, and \( \odot \) denotes the element-wise (Hadamard) product. It is important to highlight that our attention vectors act as learnable feature weightings rather than input-conditioned modules. We observe that this fixed representation with its simple design is less susceptible to forgetting, similar to the behavior of our weight component keys.

\subsection{Expansion \& Orthogonality}
\label{expand}
The key to alleviating CF is to prevent overwriting the knowledge acquired in previous tasks. When encountering a new task, we freeze the existing components and expand the collection by updating only the new components. This is illustrated at the bottom of Figure \ref{model}, where the existing parameters are \textit{frozen}, and only the newly added parameters are \textit{trainable}. Specifically, for task \( \mathcal{T}_t \), we learn \( \frac{M}{N} \) components, where \( N \) denotes the number of tasks and \( M \) is the hyperparameter, while the previously learned \( \frac{(t-1) \cdot M}{N} \) components are kept frozen. This extension is achieved through our attention-based component weighting scheme, ensuring that the expansion of parameters does not affect the computation of the weights \( \boldsymbol{\alpha} \) corresponding to the previously learned components.

To further mitigate CF, we introduce orthogonality constraints for \( \mathbf{W} \), \( \mathbf{K} \), and \( \mathbf{A} \). The underlying intuition is that interference is minimized when vectors are orthogonal. For instance, we aim to prevent the keys and weights learned in task \( \mathcal{T}_2 \) from influencing data from task \( \mathcal{T}_1\). To achieve this, we initialize the vectors orthogonally and incorporate a simple orthogonal penalty loss as:
\begin{equation}
    \mathcal{L}_{\text{ortho}}(B) = \| BB^\top - I \|_2^2
\end{equation}
where \( B \) is an arbitrary matrix, \( I \) is identity matrix, and \( \| \cdot \|_2 \) denotes Frobenius norm, which measures the squared sum of all entries of the matrix. 

Additionally, we propose a stochastic restoration method to restore knowledge from the source pre-trained model, thereby enhancing plasticity.
The update of the weight $W$ at step $t$ is as follows:
\begin{equation}
    Mask \sim \text{Bernoulli}(p)
\end{equation}
\begin{equation}
    W_{t+1} = Mask \odot W_0 + (1 - Mask) \odot W_{t+1}
\end{equation}
where \( p \) is a small restoration probability, and \( Mask \) is a mask tensor of the same shape as \( W_{t+1} \). The mask determines which elements within \( W_{t+1} \) will be restored to the source weights \( W_0 \).


\subsection{Full Optimization}
Combining all objectives, our full optimization is:
\begin{equation}
    \begin{split}
    &\min_{\mathbf{W}_n, \mathbf{K}_n, \mathbf{A}_n} \mathcal{L}_{\text{CL}}(f_{\theta, \mathbf{W}, \mathbf{K}, \mathbf{A}, \boldsymbol{\lambda}_h, \boldsymbol{\lambda}_l}(\boldsymbol{x}), \boldsymbol{y})~ + \\
    &\beta (\mathcal{L}_{\text{ortho}}(\mathbf{W}) + \mathcal{L}_{\text{ortho}}(\mathbf{K}) + \mathcal{L}_{\text{ortho}}(\mathbf{A}))
    \end{split}
\end{equation}
where \( \mathbf{W}_n, \mathbf{K}_n, \mathbf{A}_n \) refer to the weight components and corresponding keys/attention vectors that are unfrozen and trained during task $\mathcal{T}_n$ and $\beta$ is a hyperparameter balancing the orthogonality loss.

\section{Experiment}

In Section \ref{main}, we compare our method with other SOTA methods on three public benchmarks. In Section \ref{gen}, we further evaluate the task generalization ability of the proposed method on unseen target tasks. Comprehensive ablation studies are conducted in Section \ref{ab}. 

\begin{figure*}[t]
\centerline{\includegraphics[width=0.95\textwidth]{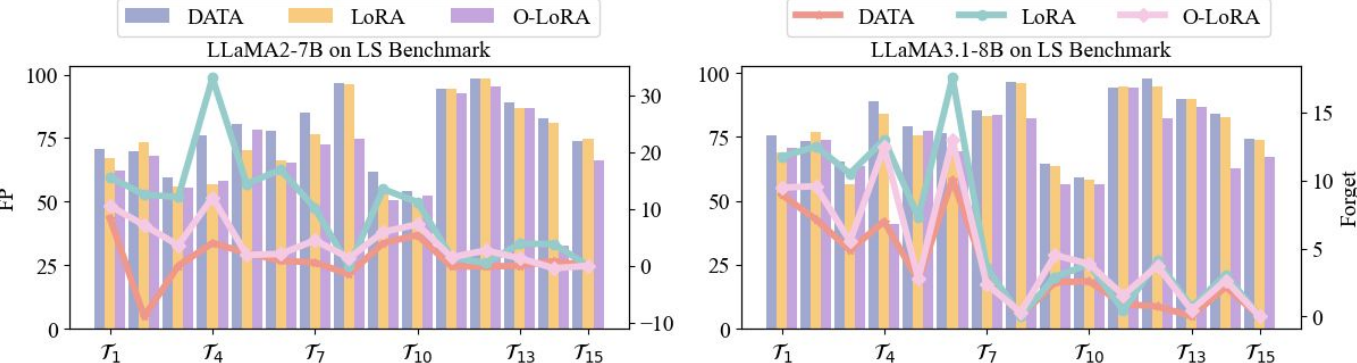}}
\caption{The shifts in CL methods with \textbf{FP} and \textbf{Forget} on LS Order 4. DATA prevents the shift (blue bar) and thus mitigates forgetting (orange line).}
\label{fig3}
\end{figure*}

\subsection{Datasets}
\label{sec:dataset}
\noindent \textbf{Standard CL Benchmark} (SC) is a CL benchmark for language models, which consists of five text classification datasets introduced by \cite{cl}. We follow \citet{olora} to pick four datasets (AG News, Amazon reviews, DBpedia and Yahoo Answers) and shuffle the tasks into three different orders to form orders 1, 2, and 3.

\noindent \textbf{Long Sequence Benchmark} (LS) is an extended version of the standard CL benchmark with 15 datasets (five classification tasks, nine GLUE and SuperGLUE tasks, and the IMDB dataset) \cite{pro}. Following \citet{pro}, we select 1,000 random samples for training each task and hold out 500 samples per class for validation and testing. Similarly, we shuffle them to form orders 4, 5, and 6.

\noindent \textbf{TRACE} is a CL benchmark for LLMs that includes 8 datasets that cover multichoice QA, multilingual capabilities, code generation, and mathematical reasoning \cite{trace}.


\noindent \textbf{Metrics.} We adopt the following three metrics to quantify various performances: 1) $\mathbf{FP} = \frac{1}{N} \sum_{j=1}^{N} a^{T_j}_N$ is the average zero-shot performance across all \( N \) tasks after tuning on the final \( N \)-th task. Here, \( a^q_{m} \) denotes the zero-shot performance on task \( q \) after sequentially tuning the \( m \)-th task, and \( T_j \) refers to the \( j \)-th task in the sequence. 2) $\mathbf{AP} = \frac{1}{N} \sum_{j=1}^{N} a_j^{T_j}$ is the average zero-shot performance when learning each \( j \)-th task, which measures the plasticity of the model. 3) $\mathbf{Forget} = \mathbf{AP} - \mathbf{FP}$ is calculated as the difference between $\mathbf{AP}$ and $\mathbf{FP}$, as commonly used in previous studies \cite{pretrained, unlocking} to quantify forgetting.
More detailed information on the datasets about orders and evaluation metrics is presented in the Appendix \ref{app_data}.

\subsection{Baselines}

We compare our method with the following baselines: 1) \textbf{SeqLoRA}: Trains fixed-size LoRA on a sequence of tasks. 
2) \textbf{LoRAReplay}: Trains new tasks on LoRA with mixing a 2\% past task. 
3) \textbf{O-LoRA} \cite{olora}: Learns tasks in different LoRA subspaces that are kept orthogonal to each other and sums all LoRA weights up at testing time. 
4) \textbf{MIGU} \cite{migu}: Only updates the model parameters with large magnitudes.

\subsection{Implementations}

We implement DATA with the LLaMA3.1-8B \cite{llama3}, LLaMA2-7B \cite{llama}, Qwen2.5-7B \cite{Qwen2TR} and T5-Large \cite{t5}. All experiments are conducted on 2 A100 GPUs with 80GB using LLaMA-Factory \cite{llamafactory}. All experimental results are reported as the average of 3 runs. Please refer to the Appendix \ref{experiment} for more detailed settings.

\subsection{Main Results}
\label{main}
To demonstrate the effectiveness of the proposed DATA method, we perform experiments on three CL benchmarks, as summarized in Table \ref{ms}. Following O-LoRA \cite{olora}, we present the results of three independent runs with different task orders on the two previous CL benchmarks. Detailed results for each order and each task within a specific order are provided in Appendix \ref{extend_result}.

\noindent \textbf{Our DATA significantly reduces the \textit{CF} of abilities and knowledge in CL.} Traditional CL approaches often result in relatively high levels of forgetting. While O-LoRA demonstrates slightly better performance yet still experiences over 12\% forgetting on the TRACE benchmark in \textbf{FP}. In contrast, DATA consistently and significantly mitigates this issue by reducing forgetting across various models and benchmarks. For instance, compared to O-LoRA, DATA increases \textbf{FP} by an average of 14.4 and decreases \textbf{Forget} by an average of 2.1 in LLaMA2. Furthermore, compared to replay methods, DATA further increases \textbf{FP} by an average of 1.3 and reduces \textbf{Forget} by an average of 1.1, highlighting its compatibility with other methods to effectively improve capabilities. DATA provides a viable solution for addressing the problem of commonsense forgetting in existing models.  

\noindent \textbf{Our DATA method does not excessively compromise the plasticity needed to learn new tasks.} Compared to other CL methods, this technique demonstrates a negligible reduction in the \textbf{FP} and \textbf{AP} metric particularly in longer sequence tasks (LS, 15 tasks) and more challenging task instruction generalizations (TRACE, 8tasks), signifying a well-maintained balance between plasticity and stability. DATA outperforms O-LoRA in LLaMA2 by an average of 12.3 in \textbf{AP}, while the difference with SeqLoRA, known for not restricting parameter updates and theoretically offering the most optimal plasticity, is less than 1. This indicates that DATA is more suitable for practical applications in LLMs.

\noindent \textbf{Longer task sequences and more challenging task instructions contribute to increased forgetting.} A comparative analysis of the \textbf{Forget} metrics for LS and TRACE with SC reveals a significant rise in forgetting. For example, in the LLaMA3.1-8B model, the decline in performance escalates from 5.8 in SC to 9.0 in LC, and further to 17.3 in TRACE. Furthermore, across all tasks, the forgetting scores \textbf{Forget} for both LLaMA2 and LLaMA3 remain below 20, demonstrating that, compared to the T5 model, LLMs exhibit a stronger capacity to mitigate CF \cite{unlocking}.

\begin{table}[t]
\centering
\renewcommand\arraystretch{1.05}
\resizebox{\linewidth}{!}{
\begin{tabular}{lcccc|cc}
\hline
\textbf{Methods} & \textbf{MMLU} & \textbf{BBH} & \textbf{GSM8K} & \textbf{AGIEval} & \textbf{FP} \\
Zero-Shot & 65.65 & 62.12 & 56.33  & 17.72 & - \\
\hdashline
SeqLoRA & 63.58 & \textbf{11.90} & 0.00 & \textbf{20.60} & 79.92\\
LoRAReplay & 60.24 & 5.99 & 1.82  & 10.69 & 80.13 \\
O-LoRA & 62.79 & 6.31 & 1.56 & 13.87 & 71.83   \\
\textbf{DATA} & \textbf{64.47} & 10.42 & \textbf{3.63} & 16.94 & \textbf{81.39}\\
\hline

\end{tabular}
}
\caption{Task generalization comparisons on unseen tasks based the LLaMA3.1-8B after training in Order 1.}
\label{table:gene}
\end{table}

\noindent \textbf{Forgetting in CL is model-dependent}. A comparison of \textbf{Forget} across different models reveals that stronger foundational models tend to mitigate CF more effectively. For instance, on the LS benchmark, SeqLoRA with LLaMA2-7B exhibits 2.7 less \textbf{Forget} than SeqLoRA with LLaMA3.1-8B and 0.8 less \textbf{Forget} than LoRAReplay. This suggests that forgetting is influenced not only by the nature of the tasks but also significantly by model-related factors such as size, architecture, and the diversity of pre-training data. 

\subsection{Performance Shifts}

In Figure \ref{fig3}, we visually depict the performance changes observed during the training process of DATA compared to other CL methods. Our proposed method clearly demonstrates its effectiveness in mitigating forgetting and enhancing plasticity. By maintaining a stable structure during the fine-tuning phase, DATA ensures consistent performance on general tasks. In sequential task transitions, it is evident that DATA demonstrates superior \textbf{FP} and reduce \textbf{Forget} compared to other methods. The line shifts further illustrate that LLaMA3.1-8B exhibits less forgetting than LLaMA2-7B, emphasizing that \textbf{forgetting is model-dependent}.

\subsection{Task Generalization}
\label{gen}

We select four benchmarks to evaluate the cross-task generalization capability of DATA, an essential dimension for assessing CL algorithms. As demonstrated in Table \ref{table:gene}, DATA effectively balances generalization and CL ability by efficiently extracting task-shared knowledge using low-rank adapters. This finding suggests that actively promoting task-shared knowledge and source capabilities between different tasks is beneficial. However, all methods experience a notable decline in performance on reasoning tasks (GSM8K and BBH), which may be attributed to the simple classification tasks potentially impairing the ability to perform generative reasoning required for instruction following.

\subsection{Ablation Study}
\label{ab}

\begin{table}[t]
\centering
\renewcommand\arraystretch{1.05}
\resizebox{\linewidth}{!}{
\begin{tabular}{cccccccc}
        \toprule
        & DATA$_h$ & DATA$_l$ & Weight & Attention & Ortho. & Rest. & \textbf{FP} $\uparrow$ \\
        \midrule
        E$_1$ & - & - & - & - & - & - & 73.0 \\
        E$_2$ & \checkmark & - & - & - & - & - & 73.7 \\
        E$_3$ & - & \checkmark & - & - & - & - & 75.4 \\
        E$_4$ & \checkmark & \checkmark & - & - & - & - & 77.6 \\
        E$_5$ & \checkmark & \checkmark  & - & - & - & \checkmark & 78.1 \\
        E$_6$ & \checkmark & \checkmark  & \checkmark & - & - & \checkmark & 78.6 \\
        E$_7$ & \checkmark & \checkmark  & \checkmark & \checkmark & - & \checkmark & 79.2 \\
        E$_8$ & \checkmark & \checkmark & \checkmark & \checkmark & \checkmark & \checkmark & \textbf{79.5} \\
        \bottomrule
    \end{tabular}
}
\caption{Ablation studies on different components.}
\label{table:ab}
\end{table}

We conduct an ablation study in Order 1 using LLaMA2-7B to evaluate the contributions of various components in our method DATA, including high-rank DATA (DATA$_h$), low-rank DATA (DATA$_l$), decomposed component weighting strategy, attention keys, orthogonality regularization (Ortho.), and stochastic restoration (Rest.). As shown in Table \ref{table:ab} (E$_2$), incorporating high-rank DATA increases the \textbf{FP} by 0.7, indicating that high-rank features can effectively extract more task-specific knowledge for adaptation to the target task. In E$_3$, low-rank DATA improves performance by 2.4 compared to E$_1$. The overall improvement in E$_4$ reaches 4.6, suggesting that both types of DATA complement each other during continual adaptation. The stochastic restoration in E$_5$ allows the model to maintain plasticity in CL, enhancing its adaptability to new tasks. E$_6$-E$_8$ achieves an additional 2 improvements, demonstrating the effectiveness of the decomposed component weighting strategy in enhancing the task representations for each DATA.

\section{Conclusion}
In this paper, we revisit existing methods for leveraging Large Language Models (LLMs) in continual learning (CL) and propose three ideal characteristics for such systems: \textit{rehearsal-free}, \textit{inference efficiency}, and \textit{plasticity}. To address the challenges associated with these characteristics, we introduce Decomposed Attention-based Task Adaptation (DATA) as a solution to catastrophic forgetting and plasticity loss. 
Moreover, we propose a \textit{decomposed component weighting} strategy to dynamically integrate the knowledge from both low-rank and high-rank DATA, thereby enhancing the unique task representation. Extensive experiments across multiple CL benchmarks and LLMs consistently validate the effectiveness of DATA.

\section*{Limitations}

\noindent \textbf{Method} The DATA method introduces a decomposed component weighting strategy and employs both high-rank and low-rank adapters, which increases the complexity of the model architecture. This complexity may lead to higher computational costs during training and inference, particularly when scaling to larger models or more tasks. Additionally, the need for dynamic weight adjustments based on task relevance and distinction may require more sophisticated optimization techniques, potentially limiting its applicability in resource-constrained environments. Furthermore, our current approach to stochastic recovery involves a step-level method, where a small proportion of parameters is recovered every 200 steps. There is significant potential to enhance this process by exploring dynamically adaptive methods that can more effectively select saturated or less important parameters for recovery. Additionally, establishing criteria to determine when recovery is necessary could optimize the process further, potentially improving model performance and efficiency.

\noindent \textbf{Task} Although DATA is designed to be rehearsal-free, it still relies on the availability of diverse and high-quality task-specific data for effective adaptation. In scenarios where task-specific data is scarce or of low quality, the method's ability to adapt and generalize may be compromised. Additionally, the method's performance on tasks with significant domain shifts or out-of-distribution data remains to be fully explored.

\noindent \textbf{Large Language Models} The effectiveness of DATA is highly dependent on the underlying LLM architecture. While the method shows promising results on models like LLaMA2, LLaMA3.1, and Qwen2.5, its performance may vary across different LLMs, especially those with significantly different architectures or pre-training objectives. Furthermore, we do not experiment with larger models like 13B and 72B due to computational or financial constraints.

\section*{Ethical Considerations}
 Our approach does not introduce ethical concerns. The datasets we used are public, and there are no privacy issues.

\section*{AI writing statement}
This paper utilized AI assistance for language polishing of the manuscript, including vocabulary correction and spell checking.

\bibliography{custom}

\appendix

\section{Overall Framework}
Drawing from the insight that LoRA \cite{hu2022lora} has exhibited superior performance, we utilize a high- and low-rank framework to ensure stability during continual task adaptation. The overall framework and the details of our method DATA are shown in Figure \ref{model}.

LoRA (Low-Rank Adaptation), as proposed by \cite{hu2022lora}, postulates that parameter changes (\(\Delta W\)) during fine-tuning occur within a low-rank subspace. This is particularly applied to the layer weights \(W_0 \in \mathbb{R}^{m \times n}\) of a model \(f_\theta\) for a downstream task. The parameter update is formulated as \(\Delta W = A \times B\), where \(A \in \mathbb{R}^{m \times r}\) and \(B \in \mathbb{R}^{r \times n}\) are two learnable matrices, and the rank \(r\) is significantly smaller than \(\min\{m, n\}\). 

For a specific layer in the model \(f_\theta\), the LoRA update is expressed as:
\[ h' = W_0x + \Delta Wx = (W_0 + AB)x \]

Here, \(h'\) represents the updated output, and \(x\) is the input to the layer. Importantly, the original weights \(W_0\) are kept frozen during the fine-tuning process, and only the matrices \(A\) and \(B\) are trainable.  

\section{Experimantal Settings}
\label{experiment}

\subsection{Datasets}
\label{app_data}

\noindent \textbf{Train Tasks.} Tables \ref{LS} and \ref{TRACE} provide detailed information on the datasets utilized in our continual learning (CL) experiments. Table \ref{LS} presents the 15 datasets included in the Long Sequence Benchmark \cite{pro}, while Table \ref{TRACE} outlines the 8 datasets from TRACE \cite{trace}. Both tables include the corresponding evaluation metrics for each dataset.

\noindent \textbf{Generalization.} We select the 1) Multitask Language Understanding (MMLU) \cite{mmlu}, which includes multiple-choice questions across 57 subjects. 2) GSM8K \cite{gsm8k}, which is a high-quality linguistically diverse multi-step elementary math reasoning dataset. 3) BIG-Bench Hard (BBH) \cite{bbh}, which includes 27 challenging tasks spanning arithmetic, symbolic reasoning, and more, derived from BIG-Bench (BB) \cite{bb}. Most of the data consists of multiple-choice questions. 4) AGIEval \cite{agieval}, which includes a wide range of high-quality official entrance exams, qualifying exams, and advanced competitions tailored to human participants.

\begin{table*}[t]
\centering
\renewcommand\arraystretch{1.05}
\resizebox{\linewidth}{!}{
\begin{tabular}{lllll}
        \hline
        \textbf{Dataset Name} & \textbf{Category} & \textbf{Task} & \textbf{Domain} & \textbf{Metric} \\
        \hline
        Yelp & CL Benchmark & Sentiment Analysis & Yelp Reviews & Accuracy \\
        Amazon & CL Benchmark & Sentiment Analysis & Amazon Reviews & Accuracy \\
        DBpedia & CL Benchmark & Topic Classification & Wikipedia & Accuracy \\
        Yahoo & CL Benchmark & Topic Classification & Yahoo Q\&A & Accuracy \\
        AG News & CL Benchmark & Topic Classification & News & Accuracy \\
        MNLI & GLUE & Natural Language Inference & Various & Accuracy \\
        QQP & GLUE & Paragraph Detection & Quora & Accuracy \\
        RTE & GLUE & Natural Language Inference & News, Wikipedia & Accuracy \\
        SST-2 & GLUE & Sentiment Analysis & Movie Reviews & Accuracy \\
        WiC & SuperGLUE & Word Sense Disambiguation & Lexical Databases & Accuracy \\
        CB & SuperGLUE & Natural Language Inference & Various & Accuracy \\
        COPA & SuperGLUE & Question and Answering & Blogs, Encyclopedia & Accuracy \\
        BoolQA & SuperGLUE & Boolean Question and Answering & Wikipedia & Accuracy \\
        MultiRC & SuperGLUE & Question and Answering & Various & Accuracy \\
        IMDB & SuperGLUE & Sentiment Analysis & Movie Reviews & Accuracy \\
        \hline
    \end{tabular}
}
\caption{The details of 15 classification datasets in the Long Sequence Benchmark \cite{pro}.  First five tasks correspond to the standard CL benchmark \cite{cl}.}
\label{LS}
\end{table*}

\begin{table*}[t]
\centering
\renewcommand\arraystretch{1.05}
\begin{tabular}{llrllr}
        \hline
        \textbf{Dataset} & \textbf{Source} & \textbf{Avg len} & \textbf{Metric} & \textbf{Language} & \textbf{\#Data} \\
        \hline
        \rowcolor{gray!20}\multicolumn{6}{l}{\textit{Domain-specific}} \\
        ScienceQA & Science & 210 & Accuracy & English & 5,000 \\
        FOMC & Finance & 51 & Accuracy & English & 5,000 \\
        MeetingBank & Meeting & 2853 & ROUGE-L & English & 5,000 \\
        \hline
        \rowcolor{gray!20}\multicolumn{6}{l}{\textit{Multi-lingual}} \\
        C-STANCE & Social media & 127 & Accuracy & Chinese & 5,000 \\
        20Minuten & News & 382 & SARI & German & 5,000 \\
        \hline
        \rowcolor{gray!20}\multicolumn{6}{l}{\textit{Code Completion}} \\
        Py150 & Github & 422 & Edim Similarity & Python & 5,000 \\
        \hline
        \rowcolor{gray!20} \multicolumn{6}{l}{\textit{Mathematical Reasoning}} \\
        NumGLUE-cm & Math & 32 & Accuracy & English & 5,000 \\
        NumGLUE-ds & Math & 21 & Accuracy & English & 5,000 \\
        \hline
    \end{tabular}
\caption{The overview of dataset statistics in TRACE \cite{trace}. The 'Source' indicates the origin of the context. 'Avg len' denotes the average length, measured in word count for English, German, and code datasets, and in character count for Chinese datasets. 'SARI' is a score that is specific to evaluating simplification tasks.}
\label{TRACE}
\end{table*}

\subsection{Task Sequence Orders}
We report task orders used for our CL experiments in Table \ref{orders}.

\begin{table*}[t]
\centering
\renewcommand\arraystretch{1.05}
\resizebox{\linewidth}{!}{
\begin{tabular}{ccl}
        \hline
        \textbf{Benchmark} & \textbf{Order}  & \textbf{Task Sequence} \\
        \hline
        \multirow{3}*{{Standard CL Becnhmark}} & 1 &  dbpedia → amazon → yahoo → ag \\
        & 2 & dbpedia → amazon → ag → yahoo \\
        & 3 &  yahoo → amazon → ag → dbpedia \\
        \hline
        \multirow{6}*{{Long Sequence Becnhmark}}& \multirow{2}*{4} & mnli → cb → wic → copa → qqp → boolqa → rte → imdb → \\
        & &  yelp → amazon → sst-2 → dbpedia → ag → multirc → yahoo \\
        &\multirow{2}*{5} & multirc → boolqa → wic → mnli → cb → copa → qqp → rte \\
        & &  → imdb → sst-2 → dbpedia → ag → yelp → amazon → yahoo \\
        & \multirow{2}*{6}& yelp → amazon → mnli → cb → copa → qqp → rte → imdb \\
        & &  → sst-2 → dbpedia → ag → yahoo → multirc → boolqa → wic \\
        \hline
        \multirow{2}*{TRACE}& \multirow{2}*{7}&  c-stance → fomc → meetingbank → py150 → scienceqa → \\
        & & numglue-cm → numglue-ds → 20minuten\\
        \hline
    \end{tabular}
}
\caption{Seven distinct orders of task sequences were employed for the experiments in continual learning. Orders 1-3 align with the Standard CL Benchmarks, as adopted in previous studies \cite{cl}. Orders 4-6 pertain to the Long Sequence Benchmarks, which encompass a total of 15 tasks \cite{pro}. Order 7 refers to the TRACE benchmark, specifically designed for LLMs, and comprises eight datasets \cite{trace}.
}
\label{orders}
\end{table*}

\begin{table}[t]
\centering
\renewcommand\arraystretch{1.05}
\resizebox{\linewidth}{!}{
\begin{tabular}{ccccc}
        \hline
        \textbf{Orders}& \textbf{2,16} & \textbf{2,8} & \textbf{4,8}& \textbf{4,16} \\
        \hline
        1 &79.4408 & 81.0921 & 80.8125 & 80.8387\\
        2 &78.5329 & 80.3684 & 80.4507 & 80.8585\\
        3 &78.9934 & 81.8882 & 80.2007 & 80.7500\\
        \hline
    \end{tabular}
}

    \caption{The performance of DATA using LLaMA3.1-8B with different high and low ranks on the standard CL benchmark.}
    \label{tab:data}
\end{table}

\subsection{Implementations}
Our implementations are based on huggingface transformers v4.45.2 \citep{transformers} using PyTorch v2.3.1 \citep{pytorch} and LlamaFactory \cite{llamafactory}. All unseen tasks generalization evaluation conducted using the Open-Compass toolkit \cite{opencompass}, adopting its default configuration.

For Standard CL Benchmark and Long Sequence Benchmark (Order 1 - Order 6), We trained the models with 1 epoch, a constant learning rate of 1e-4.

For TRACE Order 7 (C-STANCE, FOMC, MeetingBank, Py150, ScienceQA, NumGLUE-cm, NumGLUE-ds, 20Minuten), we trained with 5000 samples with a constant learning rate of 1e-4 for 5, 3, 7, 5, 3, 5, 5, 7 epochs respectively.

In a series of performance experiments, we configured various parameters as follows: the LoRA rank was set to 8 refer to Figure \ref{lora}, and the proportion of past task data mixed in LoRAReplay was set to 2\%. For the DATA model, the low-rank configuration was set to 2 and the high-rank configuration to 8. From Figure \ref{tab:data}, it can be observed that the performance of 2 and 8 is optimal. Additionally, compared to LoRA, the increase in parameters is limited, with only an additional set of LoRA with a rank of 2, achieving a balance between resources and performance. 

In terms of the decomposed component weighting strategy, we used a weight length ($L_w$) of 8. The weight component for each task was set to \(\frac{N_{\text{layer}}}{4}\), leading to a total weight calculation of \(M = \frac{N \times N_{\text{layer}}}{4}\), where \(N_{\text{layer}}\) is the number of model layers and \(N\) represents the number of tasks. The hyperparameter \(\beta\) was assigned a value of 10. For stochastic recovery, a simple strategy was applied where a small proportion of parameters was recovered every 200 training steps.  

\subsection{More Baselines}

\noindent \textbf{IncLoRA}: Incremental learning of new LoRA parameters for a sequential series of tasks (without adding any regularization or replaying samples from previous tasks).

\noindent \textbf{LFPT5} \cite{LFPT5AU}: Continuously train a soft prompt that simultaneously learns to solve tasks and generate training samples, which are subsequently used in experience replay.

\noindent \textbf{ProgPrompt} \cite{pro}: Sequentially concatenates previously learned prompts to the current one during training and testing.

\noindent \textbf{SAPT} \cite{sapt}: In the SAPT method, a Shared Attentive Learning and Selection Module (SALS) is used to guide training samples through optimal PET blocks for task-specific learning, using a unique instance-level attention mechanism. This process ensures efficient continual learning for large language models. 

\section{Extended Results}
\label{extend_result}

\subsection{Fine-grained Results for the Main  Experiments}
We report the results of each task order on the 3 benchmarks in Table \ref{extendms}. Overall, our proposed DATA demonstrates excellent capabilities in addressing CF and Loss of plasticity.

\subsection{Supplementary Motivation}
\label{sup_moti}
\begin{figure*}[t]
\centerline{\includegraphics[width=0.95\textwidth]{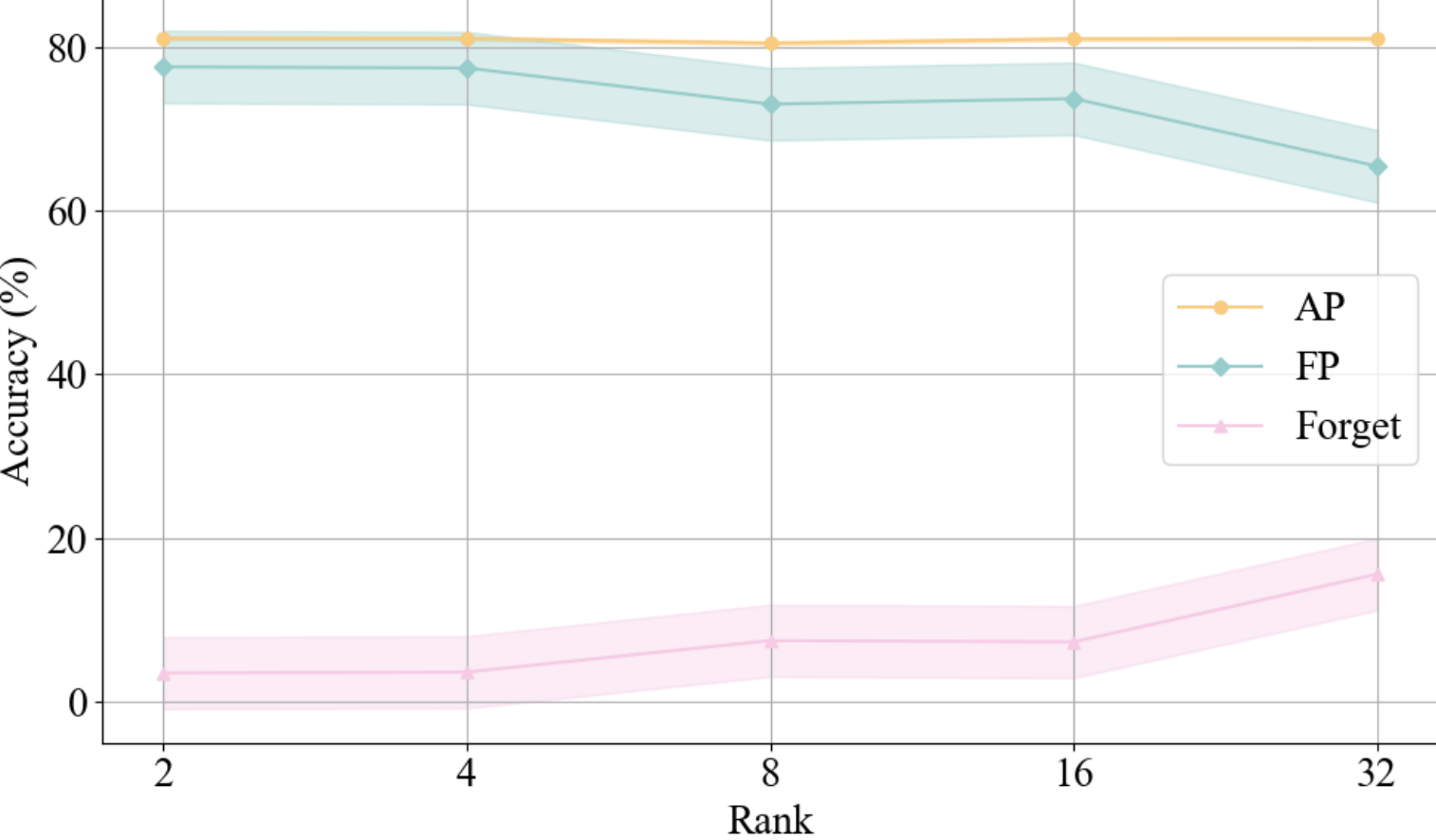}}
\caption{Performance of different LoRA ranks.}
\label{lora}
\end{figure*}

In our analysis of LoRA's performance with varying ranks on LLaMA2-7B for Order 1, as shown in Figure \ref{lora}, we observed that increasing the rank enhances Adaptation Plasticity (AP) for new tasks. However, this improvement comes at the cost of greater forgetting, as it tends to overlook the shared knowledge between tasks. Conversely, lower ranks do not lead to significant changes in forgetting, indicating that they are more effective at capturing the shared knowledge and skills across tasks. This observation suggests the necessity of integrating both high- and low-rank adapters. By doing so, we can better balance the modeling of task-specific and shared knowledge, potentially optimizing both AP and reducing forgetting.

\subsection{Efficiency Analysis}
In Table \ref{tab:comparison}, we compare the FLOPs, trainable parameters, storage features, and average predict times of various CL methods. LoRA achieves the highest inference efficiency by bypassing module selection and expansion, allowing the learned LoRA weights to be directly integrated with the original model weights during testing. Furthermore, our method utilizes only two sets of high- and low-rank adapters with a small number of parameters for decomposed component weighting and eliminates the need to store sample features. This leads to excellent performance when considering both trainable parameters and storage features.

\begin{table*}[t]
  \centering
  \resizebox{\linewidth}{!}{
      \begin{tabular}{lccccccccc}
        \toprule
          \multirow{2}{*}{\textbf{Methods}} & \multicolumn{4}{c}{\textbf{Standard CL Benchmark (SC)}}& \multicolumn{4}{c}{\textbf{Long Sequence Benchmark (LS)}} & \multicolumn{1}{c}{\textbf{TRACE}}\\
        \cmidrule(r){2-5}\cmidrule(r){6-9}\cmidrule(r){10-10}
          & \textbf{Order 1}  &\textbf{Order 2} & \textbf{Order 3} & \textbf{Avg} & \textbf{Order 4}  &\textbf{Order 5} & \textbf{Order 6} & \textbf{Avg} & \textbf{Order 7}  \\
        \midrule
        \rowcolor{gray!20} \multicolumn{10}{l}{\textit{\# T5-Large based}}\\
          SeqLoRA        & 72.1 & 66.8 & 73.3 & 70.7 & 66.4 & 63.9 & 19.5 & 59.9 & 12.1 \\
         LoRAReplay  & 74.0 & 73.1 & 73.0 & 73.3 & 74.2 & 72.7 & 73.9 &  73.6 & 34.0\\
         L2P$^*$ \cite{l2p} & 60.3 & 61.7 & 61.1 & 60.7 & 57.5 & 53.8 & 56.9 & 56.1 & -\\
         LFPT5$^*$ \cite{LFPT5AU} & 67.6 & 72.6 & \textbf{77.9} & 72.7 & 70.4 & 68.2 & 69.1 & 69.2 & -\\
         IncLoRA & 66.5 & 64.6 & 66.1 & 65.7 & 59.1 & 60.7 & 59.4 & 59.7 & -\\
         ProgPrompt$^*$ \cite{pro} & 75.2 &  75.0 & 75.1 & 75.1 & 78.0 & \textbf{77.7} & 77.9 & \textbf{77.9} & -\\
         O-LoRA \cite{olora} & 73.2 & 72.4 & 70.4 & 72.0 & 69.9 & 68.5 & 65.3 & 67.9 & - \\
         ~~~ + MIGU \cite{migu} & 73.5 & 71.4 & 70.0 & 71.6 & 65.4 & 65.2 & 65.2 & 65.3 & - \\
         SAPT-LoRA$^*$ \cite{sapt} & - & - & - & -& \textbf{83.4} & -  & \textbf{80.6} & - & - \\
         \textbf{DATA} (\textit{ours}) & 73.7 & 70.5 & 73.8 & 72.7 & 71.5 & 70.5 & 68.0 & 70.0 & 16.7\\
         ~~~ + Replay & \textbf{77.0} & \textbf{75.6} & 75.2 & \textbf{75.9} & 75.6 & 73.2 & 74.1 & 74.3 & \textbf{36.5}\\
        \midrule
        \rowcolor{gray!20} \multicolumn{10}{l}{\textit{\# LLaMA2-7B based}}\\
         SeqLoRA        & 73.0 & 73.2 & 78.4 & 74.9  & 74.7 & 73.7 & 72.5 & 73.7 & 64.1 \\
         LoRAReplay  & 80.3 & 80.4 & 76.7 & 79.2 & 80.3 & 79.5& 80.5 & 80.0 & 71.9 \\
         O-LoRA \cite{olora} & 76.2 & 76.3 & 76.8 & 76.4 & 68.5 & 67.8 & 67.5 & 67.9 & 35.0 \\
         \textbf{DATA} (\textit{ours}) & 79.5 & 79.9 & \textbf{80.0} & 79.8 & 76.6 & 77.0 & 77.2 & 76.9  & 66.0 \\
         ~~~ + Replay & \textbf{80.4} & \textbf{81.3} & 78.4 & \textbf{80.0} & \textbf{83.2} & \textbf{82.5} & \textbf{81.8} & \textbf{81.8} & \textbf{73.1}  \\
        \midrule
        \rowcolor{gray!20} \multicolumn{10}{l}{\textit{\# LLaMA3.1-8B based}} \\
         SeqLoRA        & 79.9 & 79.0 & 80.0 & 79.6 & 74.2 & 73.7 & 76.5 & 74.8 & 65.1 \\
         LoRAReplay  & 80.1  & 80.6 & 80.1 & 80.3 & \textbf{83.2} & 80.7 & \textbf{82.2}  & 82.0 & 78.7 \\
         O-LoRA \cite{olora} & 71.8 & 72.2 & 72.8 & 72.3 & 73.1 & 69.4 & 71.6 & 71.4 & 36.7 \\
         \textbf{DATA} (\textit{ours}) & \textbf{81.4} & 80.7 & 80.5 & \textbf{80.9} & 80.7 & 77.7 & 81.5 & 80.0 & 72.7 \\
         ~~~ + Replay & 80.6 & \textbf{81.0} & \textbf{80.7} & 80.8 & 83.1 & \textbf{81.7} & 81.8 & \textbf{82.2} & \textbf{80.1} \\
        \midrule
        \rowcolor{gray!20} \multicolumn{10}{l}{\textit{\# Qwen2.5-7B based}} \\
         SeqLoRA        & 80.0 & 77.9 & 78.4 & 78.8 & 79.5 & 79.1 & 81.1& 79.9 & 65.1 \\
         LoRAReplay  & \textbf{80.7}  & \textbf{80.6} & \textbf{80.1} & \textbf{80.5} & 83.3 & \textbf{83.2} & 82.7  & 83.1 & 75.7 \\
         \textbf{DATA} (\textit{ours}) & 79.8 & 79.1 & 79.4 & 79.4 & 79.8 & 80.2 & 81.5 & 80.5 & 70.4 \\
         ~~~ + Replay & 80.3 & \textbf{80.6} & 79.9 & 80.3 & \textbf{83.7} & 82.9 & \textbf{82.9} & \textbf{83.2} & \textbf{77.3} \\
        \midrule
      \end{tabular}
      }
    \caption{Summary of the results on 3 standard CL benchmarks with T5-Large, LLaMA2-7B, LLaMA3.1-8B and Qwen2.5-7B. Averaged accuracy after training on the last task (\textbf{FP}, Sec. \ref{sec:dataset}) is reported.
    }
    \label{extendms}
\end{table*}

\begin{table*}[t]
\centering
\renewcommand\arraystretch{1.05}
\resizebox{\linewidth}{!}{
\begin{tabular}{lcccc}
        \toprule
        Method & FLOPs ($10^{16}$) $\downarrow$ & Trainable Parameters (\%) $\downarrow$ & Stored Features (\%) $\downarrow$ & Predict Time (ms) $\downarrow$  \\
        \midrule
        SeqLoRA & 2.6 & 0.30 & 0 & 89 \\
        LoRARepaly & 4.8 & 0.30 & 2\% & 90  \\
        O-LoRA & 8.8 & 0.46 & 0 & 196 \\
        \textbf{DATA} & 4.6  & 0.38 & 0 & 141 \\
        \bottomrule
    \end{tabular}
}
    \caption{Comparison of the number of trainable parameters and FLOPs for Order 4 with LLaMA2-7B.}
    \label{tab:comparison}
\end{table*}

\end{document}